\documentclass{article}

\usepackage{microtype}
\usepackage{graphicx}
\usepackage{booktabs} 

\usepackage{hyperref}

\usepackage[accepted]{icml2025}

\usepackage{amsmath}
\usepackage{amssymb}
\usepackage{mathtools}
\usepackage{amsthm}

\usepackage{amsmath}
\usepackage{array}
\usepackage{multirow}
\usepackage{graphicx}
\usepackage{subcaption}
\usepackage{wrapfig}

\usepackage{inconsolata}
\usepackage{algorithm,algorithmic}

\usepackage[utf8]{inputenc} 
\usepackage[T1]{fontenc}    
\usepackage{hyperref}       
\usepackage{url}            
\usepackage{booktabs}       
\usepackage{amsfonts}       
\usepackage{nicefrac}       
\usepackage{microtype}      
\usepackage{xcolor}         
\usepackage{section}[placeins]

\usepackage{changepage}
\usepackage{adjustbox}
\usepackage{graphicx}

\newcommand{\bs}[1]{\mathbf{#1}}

\usepackage[capitalize,noabbrev]{cleveref}

\theoremstyle{plain}

\theoremstyle{definition}

\theoremstyle{remark}

\usepackage[textsize=tiny]{todonotes}

\icmltitlerunning{LCQ: Low-Rank Codebook based Quantization for Large Language Models}

\begin{document}

\twocolumn[
\icmltitle{LCQ: Low-Rank Codebook based Quantization \\ for Large Language Models}



\icmlsetsymbol{equal}{*}

\begin{icmlauthorlist}
\icmlauthor{Wen-Pu Cai}{equal,nju}
\icmlauthor{Ming-Yang Li}{equal,nju}
\icmlauthor{Wu-Jun Li}{nju}

\end{icmlauthorlist}

\icmlaffiliation{nju}{Department of Computer Science and Technology, Nanjing University, Nanjing, China}

\icmlcorrespondingauthor{Wu-Jun Li}{liwujun@nju.edu.cn}

\icmlkeywords{Model Compression, Weight Quantization}

\vskip 0.3in
]



\printAffiliationsAndNotice{\icmlEqualContribution} 

\begin{abstract}
  Large language models~(LLMs) have recently demonstrated promising performance in many tasks.
  However, the high storage and computational cost of LLMs has become a challenge for deploying LLMs. 
  Weight quantization has been widely used for model compression, which can reduce both storage and computational cost. 
  Most existing weight quantization methods for LLMs use a rank-one codebook for quantization, 
  which results in substantial accuracy loss when the compression ratio is high. 
In this paper, we propose a novel weight quantization method, called \underline{l}ow-rank \underline{c}odebook based \underline{q}uantization~(LCQ), for LLMs. 
LCQ adopts a low-rank codebook, the rank of which can be larger than one, for quantization. Experiments show that LCQ can achieve better accuracy than existing methods with a negligibly extra storage cost.
\end{abstract}

\section{Introduction}
\label{lcq:intro}

Large language models~(LLMs), such as GPT~\cite{GPT3}, GLM~\cite{GLM}, LLaMA~\cite{LLaMA} and LLaVA~\cite{LLaVA}, have recently demonstrated promising performance in many tasks including both understanding and generation of natural language, image, video and speech. However, as these models grow in size, their parameter counts also increase dramatically. For instance, the number of parameters in GPT~\cite{GPT3} can reach up to 1.8 trillion. A large number of parameters will result in high storage and computational cost, which has become a challenge for deploying LLMs in resource-constrained devices such as mobile phones or edge devices.

Weight quantization~\cite{AWQ, GPTQ, QLoRA, QA-LoRA}, which represents~(quantizes) each high bit-width weight by a low bit-width value, has been widely used for model compression. Weight quantization can reduce both storage and computational cost for model inference~\cite{GPTQ}, and hence can facilitate model deployment in real applications. For example, quantizing a model from 16-bit floating-point precision to 4-bit integers can reduce the model size by a factor of four, enabling deployment on devices with limited memory. 

Despite the benefits, naive weight quantization will typically cause accuracy degradation, especially for cases when the compression ratio is high, i.e., the bit-width after quantization is too low. Existing weight quantization methods for LLMs have proposed different techniques to mitigate the accuracy degradation caused by quantization. For example,
GPTQ~\cite{GPTQ} uses output reconstruction error rather than simple parameter reconstruction error for quantization. GPTQ~\cite{GPTQ} primarily proposes a learning method to compensate for quantization residuals and does not perform additional optimizations for the quantization codebook.
AWQ~\cite{AWQ} and OmniQuant~\cite{OmniQuant} mainly focus on learning the quantization codebook, further reducing the performance loss of the quantized model.
Although existing weight quantization methods can achieve accuracy comparable to the original model with 4-bit or larger bit-width quantization, 
there remains a significant accuracy degradation for quantization when the bit-width is lower than 4. 

One common characteristic of most existing weight quantization methods for LLMs is that they use a rank-one codebook for quantization. 
Specifically, the quantization codebook for weights can be seen as the outer product of a scaling vector~\cite{AWQ, GPTQ, QLoRA} and a quantization point set~(QPS) vector~\cite{SubsetQ}.
Each element in the scaling vector corresponds to a scaling factor for a group of weights and the QPS vector is constructed from quantiles of uniform distribution ~\cite{GPTQ, AWQ} or Gaussian distribution~\cite{QLoRA}. 
Although the rank-one codebook can reduce the storage cost of the model, it has limited representation ability. This might be one of the main reasons to explain the significant accuracy degradation for existing quantization methods when the bit-width is lower than 4.

In this paper, we propose a weight quantization method, called \underline{l}ow-rank \underline{c}odebook based \underline{q}uantization~(LCQ), for LLMs. The main contributions of LCQ are outlined as follows:
\begin{itemize}
\item LCQ adopts a low-rank codebook, the rank of which can be larger than one, for quantization. 
\item In LCQ, a gradient-based optimization algorithm is proposed to optimize the parameters of the codebook.
\item LCQ adopts a double quantization strategy for compressing the parameters of the codebook, which can reduce the storage cost of the codebook.
\item Experiments show that LCQ can achieve better accuracy than existing methods with a negligibly extra storage cost.

\end{itemize}

\section{Related Work}
\label{sec:related}
Existing weight quantization methods for LLMs typically employ post-training quantization~(PTQ) which is relatively fast and does not require access to the full original training data~\cite{GPTQ, AWQ, QLoRA}. 
PTQ methods such as BRECQ~\cite{BRECQ}, OBC~\cite{OBC} and QDrop~\cite{QDROP} are designed for quantizing traditional non-LLMs, which cannot be directly used for LLMs due to their high optimization cost. 
Recently, some PTQ methods have been specially designed for LLM quantization.
GPTQ~\cite{GPTQ} primarily proposes a learning method to compensate for quantization residuals and uses output reconstruction error as the objective function. 
However, GPTQ does not perform optimization for the quantization codebook, which results in significant accuracy loss with low bit-width. 
AWQ~\cite{AWQ} and OmniQuant~\cite{OmniQuant} mainly focus on learning the quantization codebook, and
the quantization indices are directly computed based on the index of the quantization value that is nearest to the full-precision weight.
The codebooks of AWQ and OmniQuant are rank-one, limiting the representation ability. Very recently, there appear some PTQ methods like OWQ~\cite{OWQ}, SpQR~\cite{SpQR} and SquzzeLLM~\cite{SqueezeLLM}, which  mainly address the issue of outliers in activations by 
assigning higher bit-widths to the channels/connections of activation outliers.
Our LCQ is orthogonal to methods like OWQ, SpQR and SqueezeLLM, and can be further integrated with these methods.

Moreover, some LLMs compression methods~\cite{LLMint8, SmoothQuant, RPTQ, Suppression+} integrate weight quantization and activation quantization for further accelerating LLMs inference. However, the accuracy drops significantly when quantizing both weights and activations.

PTQ methods have also been integrated with parameter-efficient fine tuning~(PEFT). Representative methods include QLoRA~\cite{QLoRA}, PEQA~\cite{PEQA}, QA-LoRA~\cite{QA-LoRA}, and LoftQ~\cite{LoftQ}. But the main goal of these methods is for PEFT rather than directly quantizing the original LLMs. 
Our LCQ can also be used for PEFT, but this is not the focus of this paper.

\section{Methodology}
As in existing weight quantization methods for LLMs~\cite{GPTQ, AWQ, OmniQuant}, we also employ post-training quantization in our LCQ.
In post-training quantization, we have a calibration dataset, which is denoted as $\{\bs{X}_i\}_{i=1}^{N_X}$. 
Here, each calibration sample $\bs{X}_i \in \mathbb{R}^{L \times D}$ has a sequence length of $L$ and feature dimension of $D$, 
and $N_X$ is the total number of calibration samples. As in AWQ~\cite{AWQ} and OmniQuant~\cite{OmniQuant}, we perform grouped quantization for the weights of linear layers. 
More specifically, for a linear layer with $N_W$ weights, we first divide all the weights into $N_V$ subsets, with each subset having $N_W/N_V$ weights. Then for each subset of weights, we group its weights into $N_G$ groups. 
We denote each subset of weights as $\bs{W} \in \mathbb{R}^{N_G \times G}$, in which $G$ denotes the weight group size and $N_G = \frac{N_W}{N_V \times G}$ denotes the number of weight groups per subset. 
A quantization codebook containing quantization values is maintained for each group of weights.
For weight quantization with $b$ bits, the number of all quantization values of each group of weights is $N_Q=2^{b}$. We denote the quantization codebook of all groups as $\bs{C} \in \mathbb{R}^{N_G \times N_Q}$, the $i$th row of which is the codebook corresponding to the $i$th group of weights.
The quantization function operates on the weights in an element-wise way, which quantizes each weight to the nearest quantization value in the codebook. Let $\bs{Z} \in [0,1,\cdots, N_Q]^{N_G \times G}$ denote the quantization index of $\bs{W}$, in which $Z_{i,j}$ denotes the quantization index of $W_{i,j}$. We have
\begin{align}\label{eq:index}
  Z_{i,j} = \mathop{\arg\min}\limits_{1 \leq k\leq N_Q} || W_{i,j}-C_{i,k}||.
\end{align}
The quantization function $Q(\cdot)$ can be defined as follows:
\begin{align}
  & Q(W_{i,j}, \bs{C}_{i}) =C_{i,Z_{i,j}}.
  \label{quantize}
\end{align}
Here, $\bs{C}_{i}$ denotes the $i$th row of $\bs{C}$. \mbox{$\bs{W}^Q = Q(\bs{W},\bs{C})$} is the quantized weight matrix with \mbox{$W_{i,j}^Q = Q(W_{i,j}, \bs{C}_{i})$}.

Then we can use the quantized weight matrix $\bs{W}^Q$ for inference. In real deployment, we only need to store the codebook matrix $\bs{C}$ and the quantization index matrix $\bs{Z}$~\cite{DeepCompress}. From~(\ref{eq:index}), we can find that $\bs{Z}$ is computed based on $\bs{C}$. 
Hence, $\bs{C}$ is the key for quantization, which will affect accuracy, storage cost and computational cost.
 
\subsection{Low-Rank Codebook}

In AWQ~\cite{AWQ} and GPTQ~\cite{GPTQ}, the intervals between quantization values are equal, and hence it is not necessary to explicitly store the quantization codebook $\bs{C}$. 
More specifically, $\bs{C}$ can be represented by two vectors, a scaling vector  $\bs{S}_1 \in \mathbb{R}^{1 \times N_G}$ and a fixed quantization point set~(QPS) vector $\bs{V}_1 \in \mathbb{R}^{1 \times N_Q}$. $\bs{V}_{1} = [-1, -1+\frac{2}{2^{b}-1}, \cdots,  1-\frac{2}{2^{b}-1}, 1]$ is a uniformly spaced vector between $[-1,1]$. 
$\bs{S}_1$ can be adaptively obtained based on the weight matrix $\bs{W}$. 
The codebook is then obtained by computing the outer product of $\bs{S}_1$ and $\bs{V}_1$. 
Furthermore, to adapt to asymmetric weight distribution, a quantization offset $\bs{B} \in \mathbb{R}^{N_G \times 1}$ is introduced, with each group of weights having its own quantization offset. 
The final quantization codebook can be formulated as $\bs{C} = \bs{S}^T_1 \bs{V}_1 - \bs{B}$, which is a rank-one matrix. The storage cost for $\bs{S}_1$, $\bs{V}_1$ and $\bs{B}$ is $O(2*N_G + N_Q)$, which is much lower than $O(N_G*N_Q)$ for storing $\bs{C}$. Although rank-one codebook can reduce the storage cost, it has limited representation ability.

\begin{figure}[t]
\vskip 0.1in
\begin{center}
\centerline{\includegraphics[width=\columnwidth]{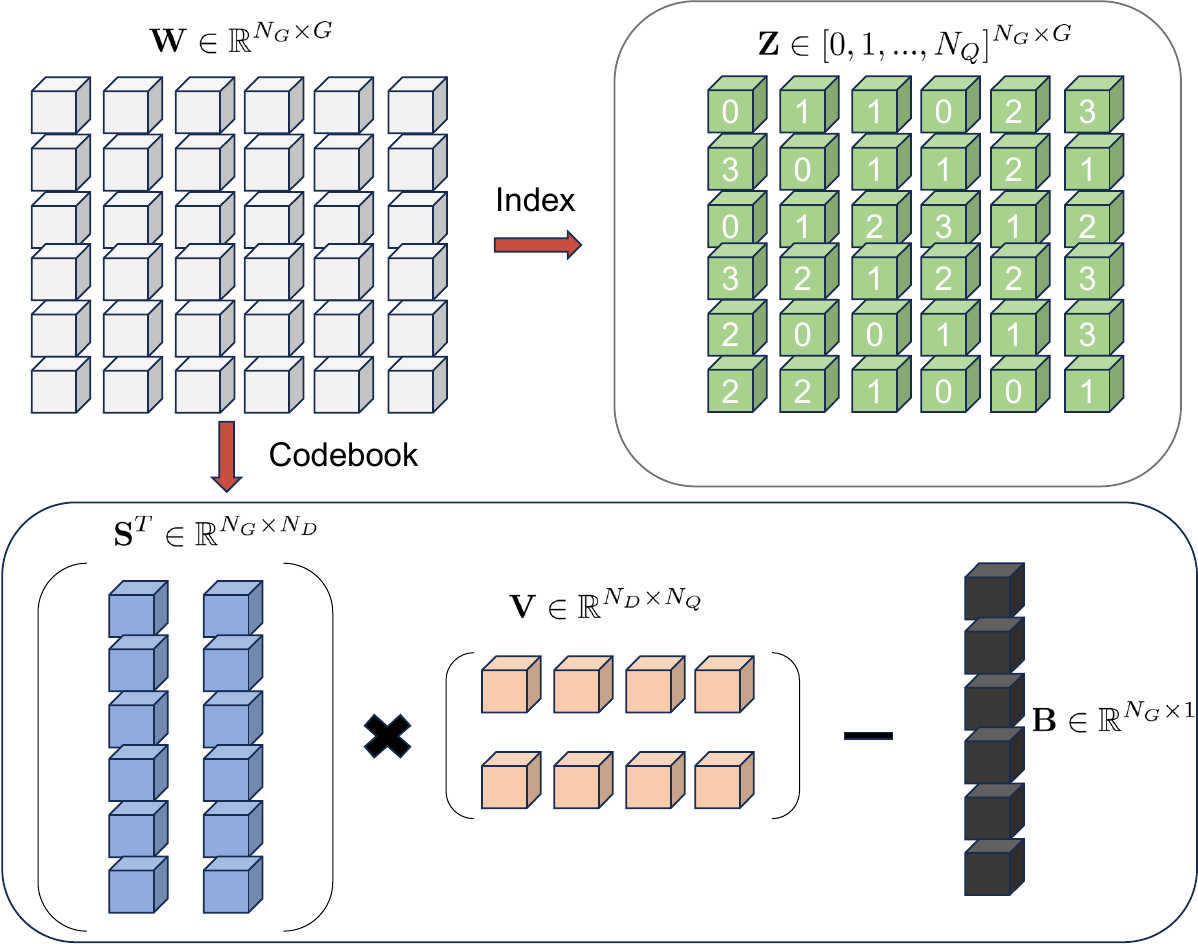}}
\caption{Illustration of low-rank codebook based quantization.}
\label{fig:LRCQ}
\end{center}
\vskip -0.1in
\end{figure}

To improve the representation ability of the codebook, we propose to use a low-rank codebook for quantization. 
More specifically, we introduce two matrices: $\bs{S} = \left[\bs{S}_1; \cdots; \bs{S}_{N_D} \right]\in \mathbb{R}^{N_D \times N_G}$, $\bs{V} = \left[\bs{V}_1; \cdots; \bs{V}_{N_D} \right] \in \mathbb{R}^{N_D \times N_Q}$.
The quantization codebook $\bs{C}$ is obtained as follows:
\begin{align}
  \bs{C} = \Phi(\bs{S}, \bs{V}, \bs{B}) = \bs{S}^T \bs{V} - \bs{B}.
\end{align}
When $N_D>1$, the rank of the codebook $\bs{C}$ can be larger than one.
When $N_D=1$, the codebook $\bs{C}$ degenerates to rank-one codebook as in existing methods~\cite{AWQ, GPTQ}.
Hence, our LCQ has a stronger representation ability than existing methods, including existing rank-one codebook based methods as special cases. 
We will show that LCQ with $N_D>1$ will achieve better accuracy than LCQ with $N_D=1$ in Section~\ref{sec:experiment}. 
This verifies the stronger representation ability of our low-rank codebook compared to the rank-one codebook. 
Figure~\ref{fig:LRCQ} shows the low-rank codebook with $N_D = 2$.

\subsection{Objective Function}
As illustrated in Figure~\ref{fig:LRCQ}, LCQ starts with a pre-trained model and tries to find the optimal $\bs{S}$, $\bs{V}$ and $\bs{B}$ for minimizing the loss caused by quantization. 
We use the output reconstruction error of a Transformer block as the objective function like the methods in~\cite{MREM, BRECQ, ZeroQuant}, 
which can make the parameters of a Transformer block be optimized simultaneously. 
Compared to the methods that use the output reconstruction error of a single linear layer as the objective function~\cite{GPTQ, AWQ}, 
using the output reconstruction error of a Transformer block can avoid falling into local optima and prevent the accumulation of quantization error. 
A Transformer block in the Transformer-based models~\cite{Transformer} typically contains the following linear layers: the query/key/value linear projection layers $\bs{W}^{qproj}, \bs{W}^{kproj}, \bs{W}^{vproj}$, the output projection layer $\bs{W}^{oproj}$ of the multi-head self-attention, and two linear layers $\bs{W}^{fc1}$ and $\bs{W}^{fc2}$ of the feed forward network. 
Therefore, the parameters of a Transformer block can be represented as $\mathcal{W} = \left[ \bs{W}^{qproj}, \bs{W}^{kproj}, \bs{W}^{vproj}, \bs{W}^{oproj}, \bs{W}^{fc1}, \bs{W}^{fc2} \right]$. 
Similarly, 
parameters $\mathcal{S}, \mathcal{V}, \mathcal{B}$ are used to represent all $\bs{S}$s, $\bs{V}$s, $\bs{B}$s within a Transformer block. 
We define the objective function to minimize the norm of the difference between the output with quantized weights and the output with unquantized weights in a Transformer block as follows:
  \begin{equation}
  \begin{split}
    \min_{\mathcal{S}, \mathcal{V}, \mathcal{B}} & \sum_{i=1}^{N_X}  ||h(\tilde{\bs{X}}_i, \mathcal{W}^Q) - h(\bs{X}^{fp}_i, \mathcal{W})||_F^2 \\ 
    & + ||h(\tilde{\bs{X}}_i, \mathcal{W}^Q) - h(\tilde{\bs{X}}_i, \mathcal{W})||_F^2.
    \label{blockmse}
  \end{split}
  \end{equation}
where $h(\cdot)$ denotes the function of a Transformer block, and $N_X$ is the number of calibration samples. 
$\bs{X}^{fp}_i$ represents the feature with all previous blocks having full precision and
$\tilde{\bs{X}}_i$ represents the feature with all previous blocks quantized.
$\mathcal{W}^Q = Q(\mathcal{W}, \Phi(\mathcal{S}, \mathcal{V}, \mathcal{B}))$ represents the quantized weights in a Transformer block.
$h(\tilde{\bs{X}}_i, \mathcal{W}^Q)$ represents the output feature of the Transformer block with all blocks quantized including this block.
The objective function contains two terms, one uses $h(\bs{X}^{fp}_i, \mathcal{W})$ as the reconstruction target and 
the other uses $h(\tilde{\bs{X}}_i, \mathcal{W})$.
Note that the quantization algorithm does not change the model weights $\bs{W}$~\cite{OmniQuant}, in order to avoid forgetting existing knowledge. 
By fixing $\bs{W}$, we can also reduce the optimization overhead by reducing the number of parameters to be optimized.

\subsection{Gradient-based Optimization}
To solve the problem in~(\ref{blockmse}), 
we propose a gradient-based optimization algorithm with strategies of gradient approximation for the quantization function, reparameterization of quantization parameters, and initialization of quantization parameters.

\subsubsection{Gradient Approximation for Quantization Function}
The $argmin(\cdot)$ operation in~(\ref{eq:index}) will make the gradient unable to back-propagate through this function.
This makes it challenging to learn the quantization codebook. 
We propose a gradient approximation strategy to address this issue.

As in N2UQ~\cite{N2UQ}, we rewrite the quantization function into a form that adds up multiple segments, 
each corresponding to the interval between two adjacent quantization values. The form is as follows:
  \begin{equation}  
  \begin{split}
     Q(W_{i,j}, \bs{C}_i)
    & = C_{i,\pi_i(1)} \\ 
    & + \sum_{k=1}^{N_Q -1} 
    \left( C_{i,\pi_i(k+1)} - C_{i, \pi_i(k)} \right) \\
    & *\chi_{0.5}^{(k)} \left(\text{clip}\left(\frac{W_{i, j} - C_{i, \pi_i(k)}}{C_{i, \pi_i(k+1)} - C_{i, \pi_i(k)}}, 0, 1\right)\right)
    \label{quantize-perm}
  \end{split}
  \end{equation}
where $\pi_i(\cdot)$ is a sorting function that is responsible for sorting each row of the quantization codebook $\bs{C}$ in ascending order. 
The function $\text{clip}(\cdot, 0, 1)$ clips values outside $[0,1]$ to be within the $[0,1]$ range.
The step function $\chi_{0.5}^{(k)}(x)$ is an element-wise function that discretizes weights, defined as follows:
  \begin{align}
    \chi_{0.5}^{(k)}(x) = 
      \begin{cases} 
      1 & \text{if } x > 0.5, \text{ or } \left( x = 0.5 \text{ and } k\%2 = 1 \right), \\
      0 & \text{if } x < 0.5, \text{ or } \left( x = 0.5 \text{ and } k\%2 = 0 \right).
      \end{cases}
    \label{Heaviside}
  \end{align}
Here, we quantize critical values that are in the middle of two quantization values to the quantization value at the even position, corresponding to the condition $k\%2 = 0$ in~(\ref{Heaviside}). 
This avoids quantizing all critical values to the smaller/larger quantization value, preventing an overall bias in the quantization results. Furthermore, the elements which make $C_{i, \pi_i(k+1)} - C_{i, \pi_i(k)}$
equal zero, i.e., two adjacent quantization values are equal, can cause division by zero. To prevent such cases, we truncate values of $C_{i, \pi_i(k+1)} - C_{i, \pi_i(k)}$ that are less than $\epsilon$ to $\epsilon$. 
We can verify that the quantization function in~(\ref{quantize-perm}) is equivalent to that in~(\ref{quantize}).

To address the issue that the gradient of the $\chi^{(k)}_{0.5}(x)$ is almost zero everywhere, we adopt the straight-through estimator~(STE) approximation for the backward pass of the step function $\chi^{(k)}_{0.5}(x)$. This is defined as follows:

\begin{align}
	\frac{\partial \chi^{(k)}_{0.5}(x)}{\partial x} =
		\begin{cases}
		1 & \text{if } 0 \leq x \leq 1,  \\
		0 & \text{otherwise}.
		\end{cases}
    \label{ste}
\end{align}

This means that the gradient for parameters within the $[0,1]$ interval is 1, while the gradient for parameters outside this interval is 0. 
The gradients for the rest of the operations in the quantization function $Q(\cdot)$ remain unchanged. By applying the chain rule, we can obtain $\frac{\partial \bs{W}^Q}{\partial \bs{S}},\frac{\partial \bs{W}^Q}{\partial \bs{V}}, \frac{\partial \bs{W}^Q}{\partial \bs{B}}$ to optimize the quantization parameters $\bs{S}, \bs{V}, \bs{B}$ with gradient descent.

\begin{figure*}[tbp]
  \vskip 0.15in
  \centering
  \includegraphics[width=0.3\textwidth]{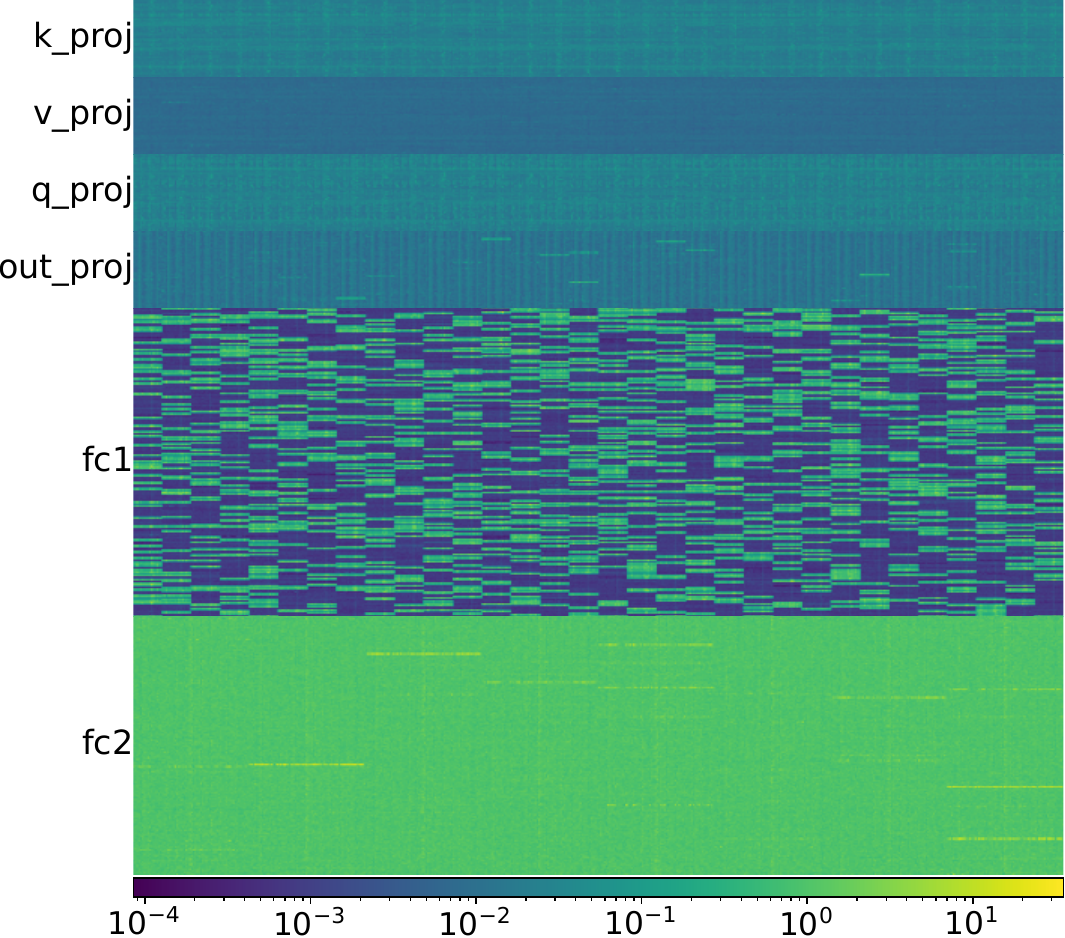}
  \includegraphics[width=0.3\textwidth]{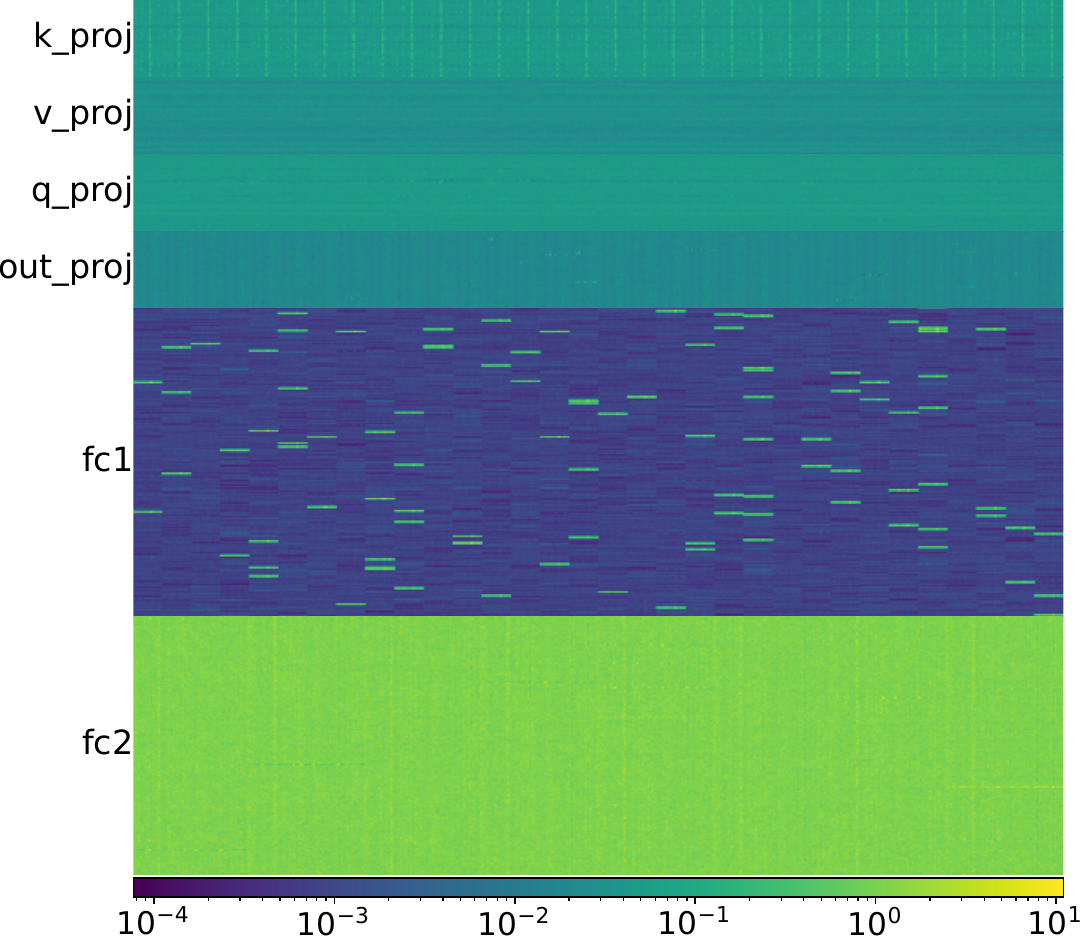}
  \includegraphics[width=0.3\textwidth]{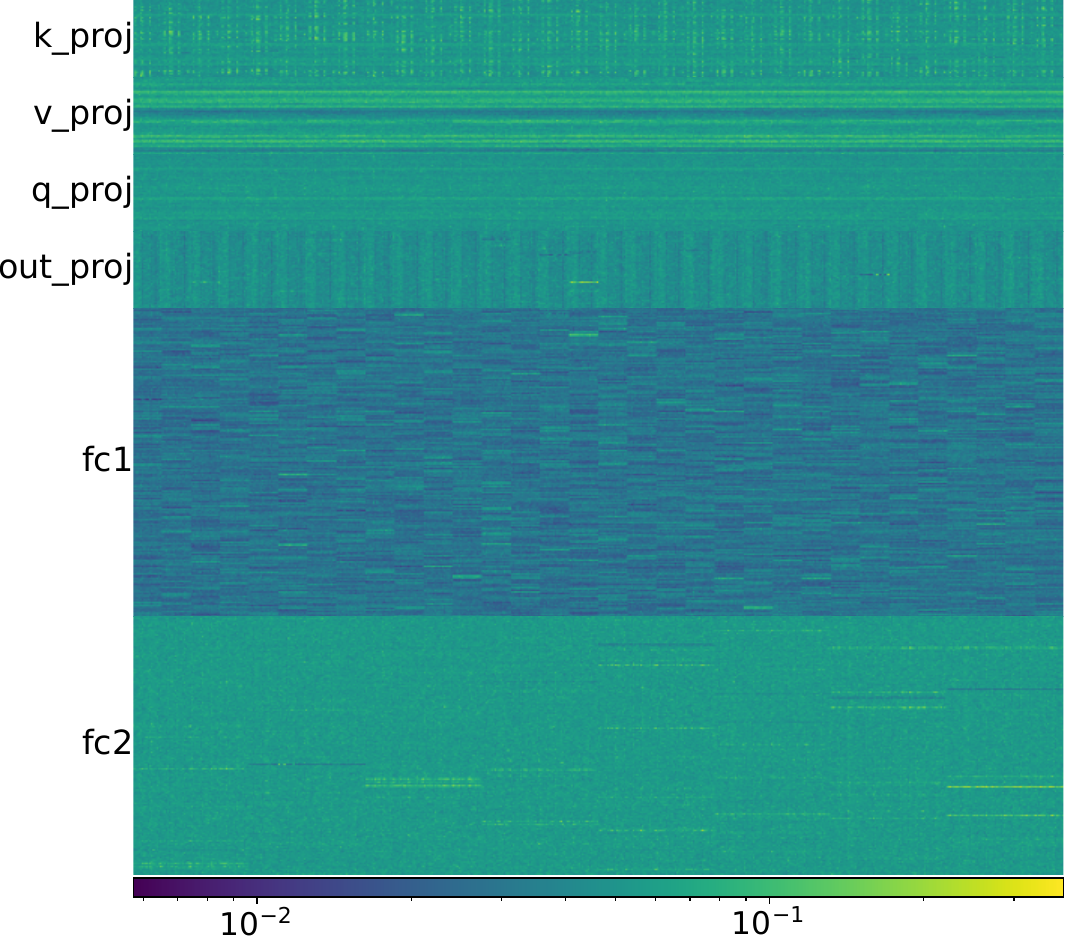}
    \caption{Visualization of $\mathcal{S}$ with AWQ, shown in the log scale. 
    3-bit quantization of a Transformer block in OPT-1.3B model.
    Left image: block 1; middle image: block 10; right image: block 20.
    }
  \label{fig:VisS}
  \vskip -0.1in
\end{figure*}

\begin{figure*}[t]
  \vskip 0.15in
  \centering  \includegraphics[width=0.85\textwidth]{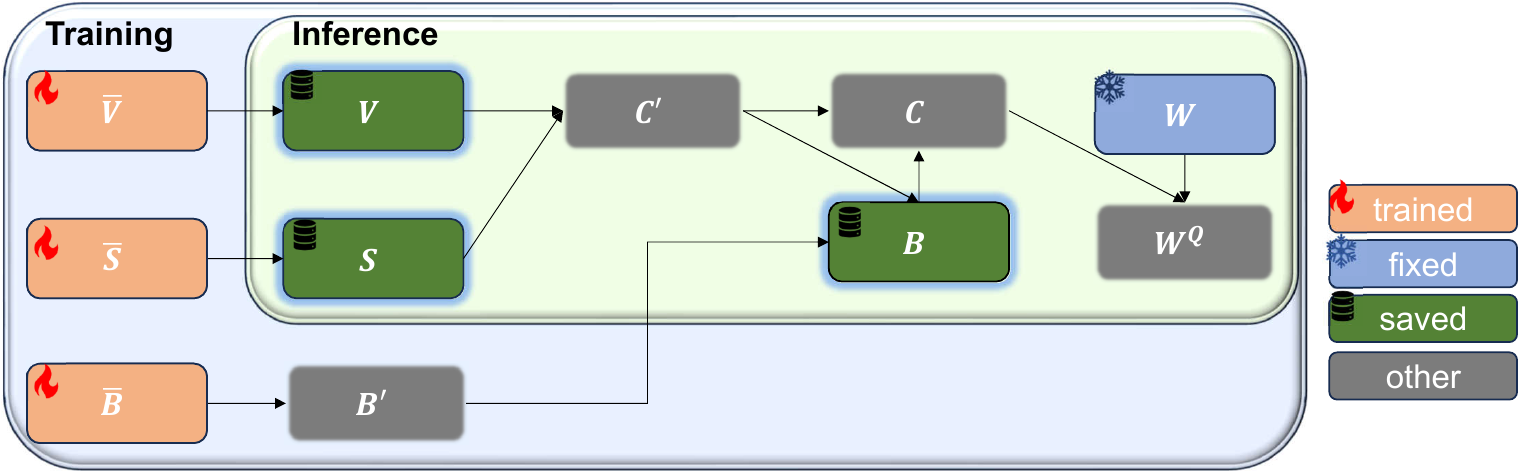}
  \caption{Training and inference process of LCQ with reparameterization.}
  \label{fig:LRCQ_tr_te}
  \vskip -0.1in
\end{figure*}

\subsubsection{Reparameterization of Quantization Parameters}
\label{sec:reparam}
During the gradient descent process, 
we find it challenging to directly learn the scaling vectors $\mathcal{S}$ for all layers in a block. 
The main reason is that the values can vary significantly among different layers, and even within the same layer~(e.g., in the fc1 layer), as shown in Figure~\ref{fig:VisS}. Directly learning $\mathcal{S}$ can result in excessive gradient updates for some weights while being too small for others, leading to difficulty in convergence. As in OmniQuant~\cite{OmniQuant}, we propose a reparameterization strategy to address the divergence issue caused by the large differences in $\mathcal{S}$ values. The reparameterization of $\bs{S}_{*, i}$ for the $i$th group of weights can be written as follows:
\begin{equation}
\begin{split}
\bs{S}_{*, i} &= g_S(\bar{\bs{S}}_{*, i}, \bs{W}_i) \\
& = \text{tanh}(\bar{\bs{S}}_{*, i}) \left( \frac{\text{max}(\bs{W}_i) - \text{min}(\bs{W}_i)}{2} \right).
\label{srep}
\end{split}
\end{equation}
Here, different groups of weights have different reparameterization coefficients $\frac{\text{max}(\bs{W}_i) - \text{min}(\bs{W}_i)}{2}$, decoupling the weight range from $\bs{S}$. After reparameterization, the parameter to be optimized changes from $\bs{S}$ to $\bar{\bs{S}}$, which is relatively uniform across different layers and can improve the consistency among gradient magnitudes to facilitate convergence. 
The use of $\text{tanh}(\cdot)$ constrains the range of $\bar{\bs{S}}$, and subsequently prevents $\bs{S}$ from exceeding the maximum range of a group of weights.

For the reparameterization of the quantization offsets $\bs{B}$, we employ a similar method as that for the reparameterization of $\bs{S}$. 
But we need to solve an extra issue. More specifically, according to previous studies~\cite{AWQ, GPTQ}, it is generally necessary to include the value $0$ in the set of quantized values. 
The main reason is that many weights in the model are close to $0$, and hence excluding $0$ from the quantized values would lead to significant quantization loss and severely degrade the accuracy of the final model. 
Assume the current offsets are $\bs{B}'$. 
For the offsets $\bs{B}'_i$ of the $i$th group, the reparameterized form is as follows:
\begin{align}
  \bs{B}_i'= g_S(\bar{\bs{B}}_{i}, \bs{W}_i).
  \label{brep}
\end{align}

A new substitute offset $\bs{B}_i$ must be recalculated in each optimization iteration based on $\bs{B}_i'$, 
ensuring that the value $0$ is included in the  codebook $\bs{C}_i$ when using $\bs{B}_i$. 
The offset calculation formula is as follows: $\bs{B}_i = Q(\bs{B}_i',\bs{C}_i')$, where $\bs{C}_i' = \bs{S}_i^T \bs{V}$. 
This ensures that $\bs{B}_i$ is definitely an element of $\bs{C}_i'$, and consequently, $0$ is assuredly an element of the quantization codebook $\bs{C}_i = \bs{S}_i^T \bs{V} - \bs{B}_i$.

We also reparameterize $\bs{V}$. As the range of weights has already been represented by $\bs{S}$, $\bs{V}$ does not need to represent the range of weights. We only need to use the $\text{tanh}(\cdot)$ function to constrain the range of $\bs{V}$ to be in $[-1,1]$:
\begin{align}
  \bs{V} = g_V(\bar{\bs{V}}) = \text{tanh}(\bar{\bs{V}}).
  \label{vrep}
\end{align}
Although the reparameterization strategy might result in a codebook in which the maximum range could exceed the maximum range of the group weights,
we can constrain the initial maximum range of the codebook to be within the maximum range of the group weights during initialization. 

Subsequent optimization during gradient descent will then fine-tune the codebook within the vicinity of the initial values. 
The training and inference process of LCQ with reparameterization is shown in Figure~\ref{fig:LRCQ_tr_te}.

\subsubsection{Initialization of Quantization Parameters}
\label{sec:initp}
We first get the initialization of $\bs{S}, \bs{V}, \bs{B}'$.
To get the initialization of reparameterization parameters, we calculate $\bar{\bs{S}}$, $\bar{\bs{V}}$, $\bar{\bs{B}}$ based on the inverse functions corresponding to~(\ref{srep}), ~(\ref{brep}), and~(\ref{vrep}).

We use the results of AWQ~\cite{AWQ} to initialize rank-1 quantization parameters. 
From the results of AWQ, we can obtain the parameters  $\bs{S}_{1}$, $\bs{V}_{1}$, $\bs{B}'$. 
Specifically, AWQ~\cite{AWQ} optimizes the equivalent transformation parameters and the clipping parameters of each linear layer.
For the equivalent transformation parameters, we apply the optimized equivalent transformation parameters of AWQ to scale the weights of each linear layer. 
For the clipping parameters, AWQ searches for the clipping parameters for each group of weights and then clips the weights.
The range and mid-point of the clipped weights correspond to $\bs{S}_{1}$, $\bs{B}'$. 
In addition, $\bs{V}_{1}$ is a vector with uniform interval, i.e., $\bs{V}_{1} = [-1, -1+\frac{2}{2^{b}-1}, \cdots, 1-\frac{2}{2^{b}-1}, 1]$ in AWQ. 

For the remaining parameters, $\bs{S}_{r}(r \geq 2)$ is initialized to zero. 
$\bs{V}_{2}$ is initialized with equidistant points from a standard Gaussian distribution and 
$\bs{V}_{r}(r \geq 3)$ is initialized with sorted random values from $\text{Uniform}(-0.1, 0.1)$.
We get the initialization of $\bar{\bs{S}}_{r}(r \geq 2)$ and $\bar{\bs{V}}_{2}(r \geq 2)$ based on the inverse functions corresponding to~(\ref{srep}) and~(\ref{vrep}).
This ensures that the initialization of LCQ corresponds exactly to the results of AWQ, 
making the optimization process easier to converge to a good result.

\subsection{Double Quantization}
\label{sec:doubleq}

When $N_D >1$, LCQ will have more parameters than rank-one codebook based methods like AWQ, which will result in extra storage cost. To mitigate the extra storage cost associated with the enlarged $\bs{S}$ and $\bs{V}$, 
we adopt a double quantization strategy as in QLoRA~\cite{QLoRA} to further quantize $\bs{S}$ and $\bs{V}$.  
Taking $\bs{S}$ as an example, with double quantization, the optimization goal is to minimize the quantization error of $\bs{S}$:
\begin{align}
	\min_{\bs{S}^{dq}, \bs{V}^{dq}, \bs{B}^{dq}} &  ||Q_{\Phi(\bs{S}^{dq}, \bs{V}^{dq}, \bs{B}^{dq})}(\bs{S}) - \bs{S}||^2_2.
  \label{dictmse}
\end{align}
Here, $\bs{S}^{dq}, \bs{V}^{dq}, \bs{B}^{dq}$ represent the parameters for double quantization. We adopt uniform quantization for double quantization. 
Like AWQ, we employ grid search to find the optimal $\bs{S}^{dq}, \bs{V}^{dq}, \bs{B}^{dq}$. 
More specifically, grid search is first used to identify the double quantization parameters for $\bs{S}$, which are then fixed. 
Subsequently, grid search is applied to determine the double quantization parameters for $\bs{V}$. 
As this process is independent of the input, the overall computational~(time) cost is negligible.

The whole optimization procedure for LCQ is outlined in Algorithm~\ref{alg:algorithm}. 
We sequentially quantize the weights $\mathcal{W}_b$ of each Transformer block.

\section{Experiments}\label{sec:experiment}

\subsection{Experimental Setup}

\begin{algorithm}[t]
  \caption{Optimization Procedure for LCQ}
  \label{alg:algorithm}
  \begin{algorithmic}
 \STATE  \textbf{Input:} Pretrained LLM, calibration dataset $\{ \bs{X}^{inp}_i \}_{i=1}^{N_X}$. \\
 \STATE  \textbf{Output:} Quantization parameters for each Transformer block: $\mathcal{S}, \mathcal{V},  \mathcal{B}$. 
  \STATE Initialize full-precision input features: $\{\bs{X}_i^{fp}\}_{i=1}^{N_X} = \{ \bs{X}^{inp}_i \}_{i=1}^{N_X}$;
  \FOR{$b = 1$ \textbf{to} $N_{block}$}
  \STATE Initialize $\bar{\mathcal{S}}, \bar{\mathcal{V}}, \bar{\mathcal{B}}$ according to Section~\ref{sec:initp};
    \FOR{$t = 1$ \textbf{to} $T$}
        \STATE Sample a mini-batch from $\{\bs{X}^{fp}_i\}_{i=1}^{N_X}$;
        \STATE Sample the same mini-batch from $\{\tilde{\bs{X}}_i\}_{i=1}^{N_X}$;
        \STATE According to Section~\ref{sec:reparam}, $\mathcal{S}, \mathcal{B}', \mathcal{V}$ are computed from $\bar{\mathcal{S}}, \bar{\mathcal{B}}, \bar{\mathcal{V}}$.  $\mathcal{B}$ is obtained from $\mathcal{B}'$;
        \STATE Compute loss according to~(\ref{blockmse}), and update parameters $\bar{\mathcal{V}}, \bar{\mathcal{S}}, \bar{\mathcal{B}}$ using gradient back-propagation with gradient approximation in~(\ref{ste});
    \ENDFOR
    \STATE Perform double quantization for $\mathcal{S}, \mathcal{V}$ according to~(\ref{dictmse}), use grid search to find the optimal double quantization codebook, and save the parameters $\mathcal{S}, \mathcal{V}, \mathcal{B}$ of the current block;
    \STATE Compute the input features with all previous blocks full-precision for the next block: $\bs{X}_i^{fp} = h(\bs{X}^{fp}_i, \mathcal{W}_b), \forall 1 \leq i \leq N_{X}$;
    \STATE Compute the input features with all previous blocks quantized for the next block: $\tilde{\bs{X}}_i = h(\tilde{\bs{X}}_i, \mathcal{W}_b^Q), \forall 1 \leq i \leq N_{X}$;
    \ENDFOR
  \end{algorithmic}
\end{algorithm}

As in AWQ~\cite{AWQ}, 128 samples randomly selected from Pile dataset~\cite{pile} are used as the calibration dataset. The sequence length of each sample is 2048.


We adopt group quantization with two settings for group size: ``G128'' denotes a group size of 128, and ``G-1'' denotes that the group size is equal to the size of the channel. $N_D$ is set to 2 by default.

\begin{table*}[t]
\center
\small
\caption{Natural language generation ability on OPT, LLaMA and LLaVA models, with perplexity on WikiText2 as the metric. The best is shown in bold. The retention rate reflects the degree of model compression.}
\vskip 0.15in
\begin{tabular}{ccccccccccc}
\hline
\multicolumn{1}{c|}{\multirow{2}{*}{}} & \multicolumn{1}{c|}{\multirow{2}{*}{fp16}} & \multicolumn{3}{c|}{W2 G128}                  & \multicolumn{3}{c|}{W3 G-1}                  & \multicolumn{3}{c}{W3 G128}        \\ \cline{3-11} 
\multicolumn{1}{c|}{}                     & \multicolumn{1}{c|}{}                      & AWQ    & Omniquant & \multicolumn{1}{c|}{LCQ} & AWQ   & Omniquant & \multicolumn{1}{c|}{LCQ} & AWQ   & Omniquant & LCQ            \\ \hline
Retention Rate                            & 1                                          & 0.134  & 0.134     & 0.138                    & 0.188 & 0.188     & 0.188                    & 0.197 & 0.197     & 0.202          \\ \hline
OPT-125M                                  & 27.65                                      & 340.56 & 117.25    & \textbf{68.75}           & 57.46 & 41.50     & \textbf{36.88}           & 35.99 & 35.23     & \textbf{32.47} \\
OPT-1.3B                                  & 14.62                                      & 42.72  & 34.69     & \textbf{25.28}           & 24.67 & 18.85     & \textbf{17.54}           & 16.41 & 16.37     & \textbf{15.97} \\
OPT-6.7B                                  & 10.86                                      & 16.73  & 17.15     & \textbf{15.49}           & 16.54 & 12.98     & \textbf{11.96}           & 11.36 & 11.53     & \textbf{11.36} \\
LLaMA-7B                                  & 5.68                                       & 12.90  & 12.33     & \textbf{9.14}            & 8.43  & 6.73      & \textbf{6.41}            & 6.31  & 6.30      & \textbf{6.17}  \\
LLaMA-13B                                 & 5.09                                       & 9.73   & 9.24      & \textbf{7.40}            & 6.37  & 5.91      & \textbf{5.60}            & 5.50  & 5.53      & \textbf{5.49}  \\
LLaVA-7B                                  & 6.83                                       & 23.63  & 36.58     & \textbf{12.29}           & 20.97 & 8.74      & \textbf{8.05}            & 7.87  & 7.86      & \textbf{7.50}  \\
LLaVA-13B                                 & 5.98                                       & 11.89  & 18.08     & \textbf{9.27}            & 8.28  & 6.99      & \textbf{6.65}            & 6.57  & 6.66      & \textbf{6.52}  \\ \hline
\end{tabular}
\label{ppl}
\vskip -0.1in
\end{table*}

\begin{table*}[t]
\center
\small
\caption{Natural language generation ability on OPT, LLaMA and LLaVA models, with  perplexity on c4 as the metric. The best is shown in bold. The retention rate reflects the degree of model compression.}
\vskip 0.15in
\begin{tabular}{ccccccccccc}
\hline
\multicolumn{1}{c|}{\multirow{2}{*}{}} & \multicolumn{1}{c|}{\multirow{2}{*}{fp16}} & \multicolumn{3}{c|}{W2 G128}                  & \multicolumn{3}{c|}{W3 G-1}                  & \multicolumn{3}{c}{W3 G128}        \\ \cline{3-11} 
\multicolumn{1}{c|}{}                  & \multicolumn{1}{c|}{}                      & AWQ    & Omniquant & \multicolumn{1}{c|}{LCQ} & AWQ   & Omniquant & \multicolumn{1}{c|}{LCQ} & AWQ   & Omniquant & LCQ            \\ \hline
Retention Rate                         & 1                                          & 0.134  & 0.134     & 0.138                    & 0.188 & 0.188     & 0.188                    & 0.197 & 0.197     & 0.202          \\ \hline
OPT-125M                               & 26.56                                      & 246.47 & 82.47     & \textbf{54.20}           & 48.83 & 36.36     & \textbf{33.81}           & 32.63 & 31.92     & \textbf{30.12} \\
OPT-1.3B                               & 16.07                                      & 40.63  & 31.59     & \textbf{25.31}           & 24.55 & 19.82     & \textbf{18.76}           & 17.80 & 17.69     & \textbf{17.30} \\
OPT-6.7B                               & 12.71                                      & 19.55  & 18.34     & \textbf{16.93}           & 17.79 & 15.20     & \textbf{13.81}           & 13.41 & 13.43     & \textbf{13.30} \\
LLaMA-7B                               & 7.34                                       & 16.55  & 14.79     & \textbf{11.44}           & 11.13 & 8.73      & \textbf{8.35}            & 8.17  & 8.18      & \textbf{8.02}  \\
LLaMA-13B                              & 6.80                                       & 12.69  & 11.60     & \textbf{9.76}            & 8.37  & 7.67      & \textbf{7.42}            & 7.34  & 7.35      & \textbf{7.26}  \\
LLaVA-7B                               & 9.27                                       & 27.83  & 33.54     & \textbf{14.74}           & 24.66 & 12.11     & \textbf{11.04}           & 10.70 & 10.79     & \textbf{10.32} \\
LLaVA-13B                              & 8.26                                       & 16.42  & 20.35     & \textbf{12.02}           & 11.46 & 9.84      & \textbf{9.37}            & 9.12  & 9.26      & \textbf{9.03}  \\ \hline
\end{tabular}
\label{c4ppl}
\vskip -0.1in
\end{table*}

We perform experiments with PyTorch~\cite{PyTorch} on an RTX A6000 GPU card of 48GB memory, and use mixed-precision optimization to reduce the memory and computational cost during training. 
The ADAMW optimizer~\cite{ADAMW} is used, with an initial learning rate set to 0.01, and the weight decay factor set to 0. 
The learning rate is gradually reduced using a cosine annealing schedule~\cite{Cosine} over a total of 10 training epochs. 
The batch size for training is set to 4, and gradient accumulation is employed to prevent memory overflow.
The weights of linear layer in all Transformer blocks are quantzied. The embedding layer and the output layer are not quantized. 

\subsection{Comparison with Baselines}

We choose AWQ~\cite{AWQ} and {OmniQuant}~\cite{OmniQuant} as baselines for comparison, because they are the most related works and have achieved state-of-the-art performance.

We first compare LCQ with baselines on the task of natural language generation, using OPT~\cite{OPT}, LLaMA~\cite{LLaMA} and LLaVA~\cite{LLaVA} as pre-trained models. 
As in AWQ~\cite{AWQ}, we use datasets WikiText2~\cite{wikitext} and c4~\cite{C4}, with the perplexity~(ppl) metric for evaluating the ability of the model to predict the next word~(i.e., generative capability). 
The \emph{retention rate}, defined as the rate between the remaining parameters to the original parameters, reflects the degree of model compression. The higher the retention rate is, the lower the degree of model compression will be. 
Table~\ref{ppl} shows the perplexity 
 on \mbox{WikiText2} dataset, and Table~\ref{c4ppl} shows the perplexity on c4 dataset. ``fp16'' denotes the original non-compressed model with 16-bit full-precision representation. ``W2 G128'' denotes a 2-bit weight quantization with a group size of 128. 
Other notations are defined similarly. 
We can find that LCQ achieves better accuracy than the baselines. 
In particular, in the case of 2-bit quantization, LCQ achieves significantly better accuracy than baselines. This significant improvement in accuracy demonstrates the effectiveness of the low-rank codebook in capturing the nuances of language modeling tasks.
Furthermore, LCQ has a comparable retention rate with baselines, which means that the increase of $N_D$ brings a negligibly extra storage cost.

\begin{figure*}[t]
  \vskip 0.15in
  \centering
  \begin{subfigure}{0.24\textwidth}
      \includegraphics[width=\textwidth,height=3.5cm]{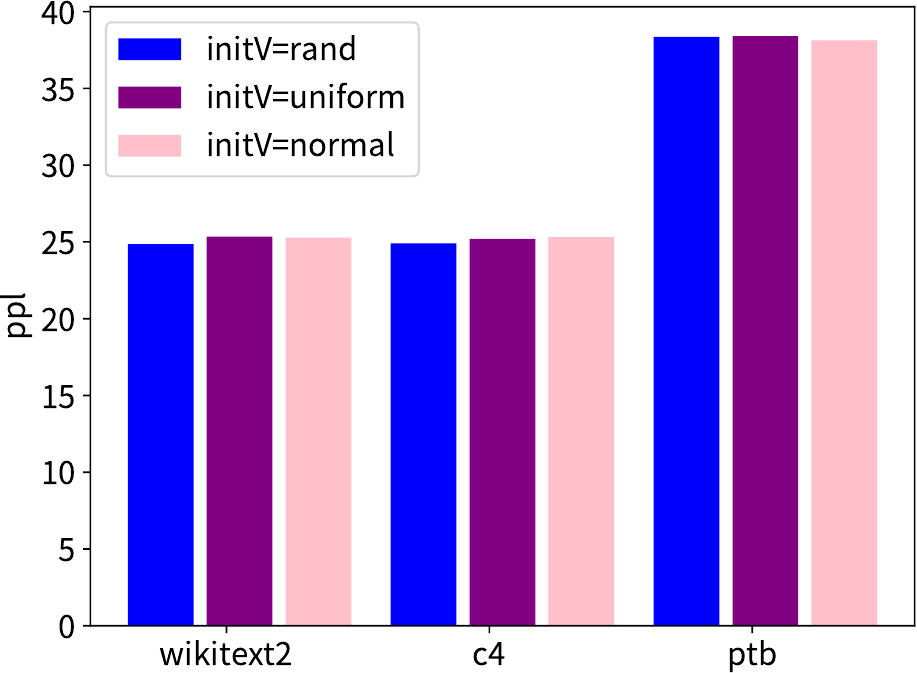} 
      \caption{Initialization for $\bs{V}$.}
      \label{fig:initV}
  \end{subfigure}
  \begin{subfigure}{0.24\textwidth}
    \includegraphics[width=\textwidth,height=3.5cm]{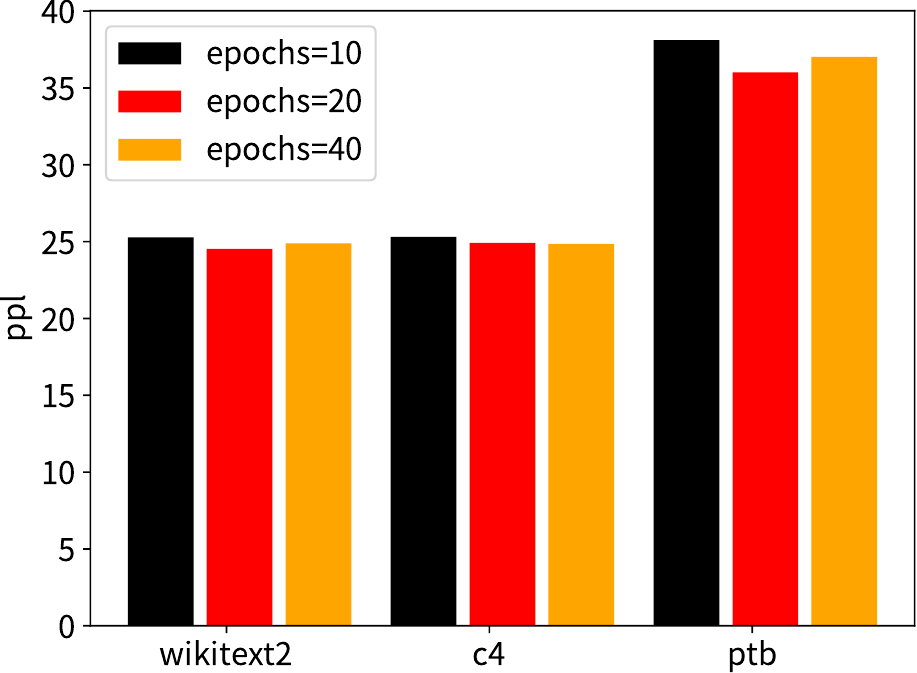}
      \caption{Training epochs.}
      \label{fig:epochs}
  \end{subfigure}
  \begin{subfigure}{0.25\textwidth}
    \includegraphics[width=\textwidth,height=3.59cm]{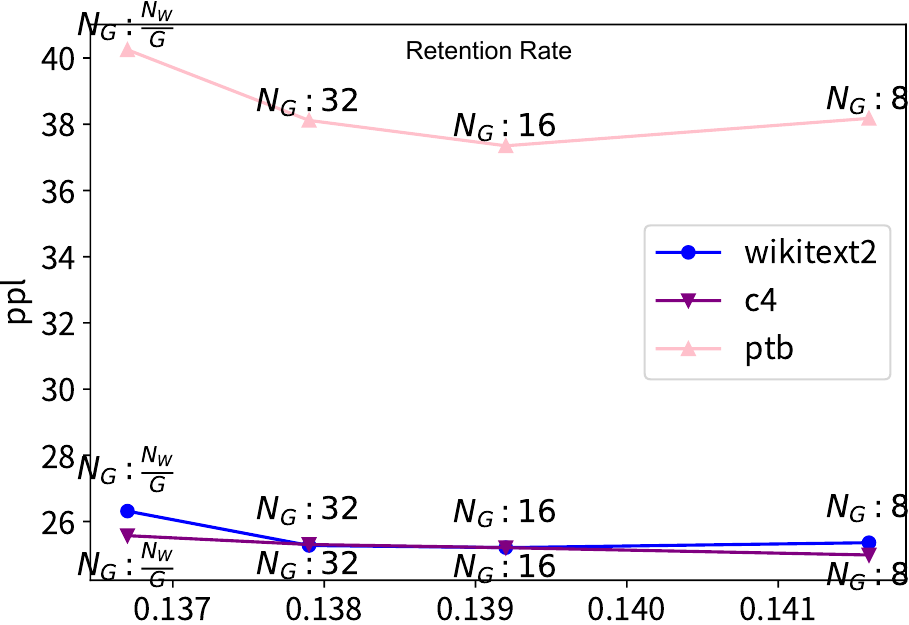} 
      \caption{$N_{G}$.}
      \label{fig:NG}
  \end{subfigure}
  \begin{subfigure}{0.25\textwidth}
    \includegraphics[width=\textwidth,height=3.48cm]{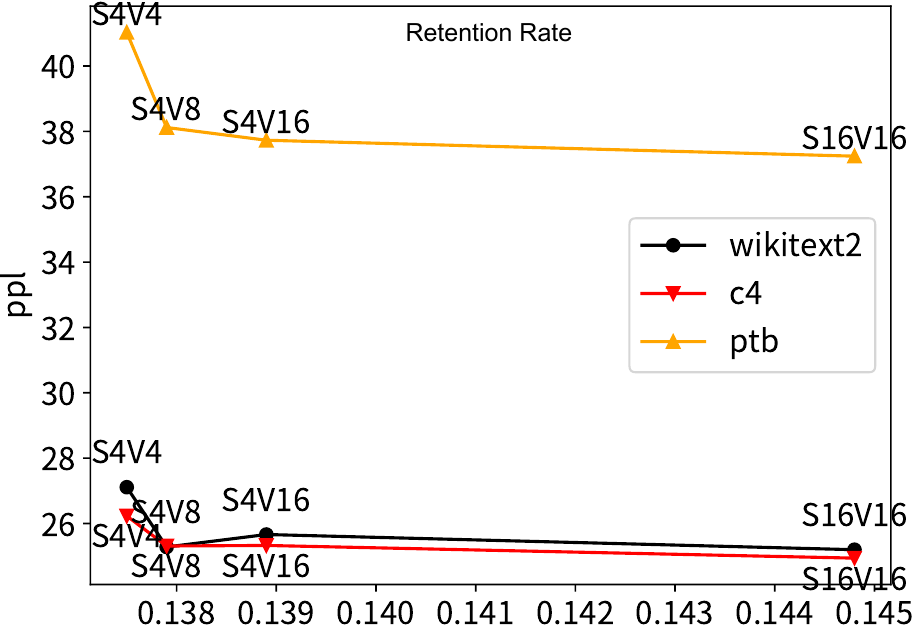} 
      \caption{Double quantization.}
      \label{fig:DQ}
  \end{subfigure}
  \caption{Hyperparameter sensitivity with ``W2 G128'' for OPT-1.3B.}
  \label{fig:hyper}
  \vskip -0.15in
\end{figure*}

\begin{table}[t]
\begin{small}
\caption{Zero-shot performance of the quantized model for LLaMA-7B, with accuracy as the metric. The best is shown in bold. ARC denotes ARC\_easy.}
\vskip 0.1in
\begin{tabular}{|c|c|c|c|c|}
\hline
		Config & Method                      & PIQA   & WinoGrande & ARC \\
	\hline
    fp16          & -                           & 78.40  & 66.93      & 67.34 \\  
    \hline
    \multirow{3}{*}{W2 G128}  
    & AWQ              & 69.21  & 55.80      & 52.74  \\
    & OmniQuant   & 66.92  & 53.20      & 49.37  \\
    & LCQ                          & \textbf{72.36} & \textbf{60.30} & \textbf{59.13} \\
    \hline
		\multirow{3}{*}{W3 G128}  
    & AWQ            & 76.33  & 65.19      & 65.07  \\
    & OmniQuant  & 76.33  & 64.96      & 64.44  \\
    & LCQ                          & \textbf{77.31} & \textbf{66.38} & \textbf{67.13} \\
		\hline
\end{tabular}
\label{Zeroshot}
\end{small}
\vskip -0.1in
\end{table}%

In addition to perplexity, we also compare LCQ with baselines on zero-shot tasks using three widely used datasets PIQA~\cite{PIQA}, WinoGrande~\cite{WinoGrande} and ARC\_easy~\cite{ARC}. Table~\ref{Zeroshot} shows the accuracy of the quantized model for LLaMA-7B. We can find that LCQ achieves better accuracy than baselines on zero-shot tasks for all datasets. In particular, in the case of 2-bit quantization, the improvement is significant compared to baselines. 
The performance on zero-shot tasks also verifies that our LCQ method does not over-fit the calibration dataset and can maintain the generalization capability of the original model on zero-shot tasks.

\subsection{Hyperparameter Sensitivity}

First, we investigate the impact of different initialization methods for LCQ. 
We only perform experiments about the initialization for $\bs{V}_{2}$, and fix the initialization of other quantization parameters as described in Section~\ref{sec:initp}. 
We compare three initialization methods: rand, uniform and normal. 
\emph{rand} adopts sorted random values from $\text{Uniform}(-0.1, 0.1)$ for initialization.
\emph{uniform} adopts the equidistant quantiles of a uniform distribution $\text{Uniform}(-1, 1)$ for initialization. 
\emph{normal} adopts the equidistant quantiles of a Gaussian distribution $\mathcal{N}(0, 1)$ ~\cite{QLoRA} for initialization. 
Figure~\ref{fig:initV} shows the results. We can find that different initialization methods do not have a significant impact on the performance. Hence, LCQ is robust to the initialization methods for $\bs{V}_{2}$. 
By default, we use \emph{normal} for initialization to avoid extra hyperparameter settings.

\begin{table}[t]
\centering
\small
\caption{Perplexity on WikiText2 when using  different $N_D$ and whether fixing $\bs{S}_1, \bs{B}, \bs{V}_1$, with ``W2 G128'' for OPT-1.3B.}
\vskip 0.1in
\begin{tabular}{|c|c|c|c|c|c|}
		\hline
		        $N_D$      & Retention Rate & Fix  & OPT-125M       & OPT-1.3B  \\
		\hline
            1        & 0.135 & $\surd$                          & 340.56         & 42.72 \\
            1        & 0.135 & $\times$                         & 88.89          & 28.50 \\
                  \hline
						2         & 0.137 & $\surd$                         & 76.98          & 27.09 \\
		        2         & 0.138 & $\times$                        & 68.75          & 25.28  \\
                  \hline
            3         & 0.141 & $\surd$                          & 65.52          & 25.15 \\
            3         & 0.141 & $\times$                         & \textbf{61.49} & \textbf{24.82}  \\
		\hline
	\end{tabular}
  \label{hyper:rank}
\vskip -0.1in
\end{table}

We also study the impact of the number of training epochs. 
Figure~\ref{fig:epochs} shows the result. We can find that 10 training epochs can achieve results comparable to those of 40 epochs. Hence, to expedite the training process, we adopt 10 training epochs by default.

The result about $N_G$ is in Figure~\ref{fig:NG}. 
The smaller the $N_G$ is, the finer the quantization granularity will be. 
In particular, $N_G=\frac{N_W}{G}$ corresponds to the coarsest granularity, 
indicating that the whole linear layer corresponds a single $\bs{V}$.
We can find that reducing $N_{G}$ can bring a slight improvement in model accuracy but also introduces some additional storage cost. 
To balance model accuracy and storage cost, we set $N_{G}=32$ by default.

The number of bits for double quantization will also affect the performance. Figure~\ref{fig:DQ} shows the result. 
In particular, $\bs{S}_1$ is not quantized and kept in fp16, consistent with baseline methods.
We can find that when using 4-bit quantization for $\bs{S}$ and 8-bit quantization for $\bs{V}$, the accuracy loss is minor or even negligible compared to methods without double quantization. However, when the number of quantization bits for $\bs{V}$ is reduced to 4, the accuracy loss is more significant. Hence, we use 4-bit quantization for $\bs{S}$ and 8-bit quantization for $\bs{V}$ by default. Double quantization is performed in groups of 16 elements by default.

We also explore the impact of $N_D$ on model performance, the result of which is shown in Table~\ref{hyper:rank}. We can find that compared to the results with $N_D=1$, 
the results with $N_D=2$ improve significantly, which verifies that increasing the rank of the codebook can lead to better accuracy. 
Further increasing $N_D$ can further improve the accuracy, 
but will also incur more storage cost. Hence, we set $N_D$ to 2 by default. Furthermore, we can also find that although the optimized parameters of AWQ are used to initialize $\bs{S}_1, \bs{V}_1, \bs{B}$ in LCQ, further learning~(optimizing) these parameters can achieve better accuracy than fixing them.

\section{Conclusion}
In this paper, we propose a novel weight quantization method, called low-rank codebook based quantization~(LCQ), for large language models (LLMs). LCQ leverages a low-rank codebook with a rank potentially greater than one, significantly enhancing the accuracy of quantized models while incurring only a negligible additional storage cost.
Experiments demonstrate that LCQ surpasses existing methods, especially in low bit-width scenarios. It effectively reduces storage and computational requirements while maintaining high accuracy, making LCQ a practical solution for deploying large models on resource-constrained devices.

\section*{Impact Statement}
The LCQ method represents a significant advancement in the field of model compression, offering a practical solution to the challenges of deploying large language models in resource-constrained environments. By enabling high compression ratios with minimal accuracy loss, LCQ makes it feasible to deploy LLMs on mobile and edge devices, opening up new possibilities for real-time applications in natural language processing and multimodal tasks. Furthermore, the reduced energy consumption associated with LCQ contributes to the development of sustainable AI technologies, addressing the growing environmental concerns related to large-scale AI model deployment. The compatibility of LCQ with existing quantization techniques ensures its versatility and broad applicability, making it a valuable tool for both academia and industry.

\bibliography{example_paper}

\begin{thebibliography}{37}
\providecommand{\natexlab}[1]{#1}
\providecommand{\url}[1]{\texttt{#1}}
\expandafter\ifx\csname urlstyle\endcsname\relax
  \providecommand{\doi}[1]{doi: #1}\else
  \providecommand{\doi}{doi: \begingroup \urlstyle{rm}\Url}\fi

\bibitem[Bai et~al.(2022)Bai, Hou, Shang, Jiang, King, and Lyu]{MREM}
Bai, H., Hou, L., Shang, L., Jiang, X., King, I., and Lyu, M.~R.
\newblock Towards efficient post-training quantization of pre-trained language
  models.
\newblock In \emph{NeurIPS}, pp.\  1405--1418, 2022.

\bibitem[Bisk et~al.(2020)Bisk, Zellers, Bras, Gao, and Choi]{PIQA}
Bisk, Y., Zellers, R., Bras, R.~L., Gao, J., and Choi, Y.
\newblock {PIQA:} reasoning about physical commonsense in natural language.
\newblock In \emph{AAAI}, pp.\  7432--7439, 2020.

\bibitem[Brown et~al.(2020)Brown, Mann, Ryder, Subbiah, Kaplan, Dhariwal,
  Neelakantan, Shyam, Sastry, Askell, Agarwal, Herbert{-}Voss, Krueger,
  Henighan, Child, Ramesh, Ziegler, Wu, Winter, Hesse, Chen, Sigler, Litwin,
  Gray, Chess, Clark, Berner, McCandlish, Radford, Sutskever, and Amodei]{GPT3}
Brown, T.~B., Mann, B., Ryder, N., Subbiah, M., Kaplan, J., Dhariwal, P.,
  Neelakantan, A., Shyam, P., Sastry, G., Askell, A., Agarwal, S.,
  Herbert{-}Voss, A., Krueger, G., Henighan, T., Child, R., Ramesh, A.,
  Ziegler, D.~M., Wu, J., Winter, C., Hesse, C., Chen, M., Sigler, E., Litwin,
  M., Gray, S., Chess, B., Clark, J., Berner, C., McCandlish, S., Radford, A.,
  Sutskever, I., and Amodei, D.
\newblock Language models are few-shot learners.
\newblock In \emph{NeurIPS}, pp.\  1877--1901, 2020.

\bibitem[Clark et~al.(2018)Clark, Cowhey, Etzioni, Khot, Sabharwal, Schoenick,
  and Tafjord]{ARC}
Clark, P., Cowhey, I., Etzioni, O., Khot, T., Sabharwal, A., Schoenick, C., and
  Tafjord, O.
\newblock Think you have solved question answering? try arc, the {AI2}
  reasoning challenge.
\newblock \emph{CoRR}, abs/1803.05457, 2018.

\bibitem[Dettmers et~al.(2022)Dettmers, Lewis, Belkada, and
  Zettlemoyer]{LLMint8}
Dettmers, T., Lewis, M., Belkada, Y., and Zettlemoyer, L.
\newblock Llm.int8(): 8-bit matrix multiplication for transformers at scale.
\newblock In \emph{{NeurIPS}}, pp.\  30318--30332, 2022.

\bibitem[Dettmers et~al.(2023)Dettmers, Pagnoni, Holtzman, and
  Zettlemoyer]{QLoRA}
Dettmers, T., Pagnoni, A., Holtzman, A., and Zettlemoyer, L.
\newblock Qlora: Efficient finetuning of quantized llms.
\newblock In \emph{NeurIPS}, pp.\  10088--10115, 2023.

\bibitem[Dettmers et~al.(2024)Dettmers, Svirschevski, Egiazarian, Kuznedelev,
  Frantar, Ashkboos, Borzunov, Hoefler, and Alistarh]{SpQR}
Dettmers, T., Svirschevski, R., Egiazarian, V., Kuznedelev, D., Frantar, E.,
  Ashkboos, S., Borzunov, A., Hoefler, T., and Alistarh, D.
\newblock Spqr: {A} sparse-quantized representation for near-lossless {LLM}
  weight compression.
\newblock In \emph{ICLR}, 2024.

\bibitem[Frantar \& Alistarh(2022)Frantar and Alistarh]{OBC}
Frantar, E. and Alistarh, D.
\newblock Optimal brain compression: {A} framework for accurate post-training
  quantization and pruning.
\newblock In \emph{NeurIPS}, pp.\  4475--4488, 2022.

\bibitem[Frantar et~al.(2023)Frantar, Ashkboos, Hoefler, and Alistarh]{GPTQ}
Frantar, E., Ashkboos, S., Hoefler, T., and Alistarh, D.
\newblock {GPTQ:} accurate post-training quantization for generative
  pre-trained transformers.
\newblock In \emph{{ICLR}}, 2023.

\bibitem[Gao et~al.(2021)Gao, Biderman, Black, Golding, Hoppe, Foster, Phang,
  He, Thite, Nabeshima, Presser, and Leahy]{pile}
Gao, L., Biderman, S., Black, S., Golding, L., Hoppe, T., Foster, C., Phang,
  J., He, H., Thite, A., Nabeshima, N., Presser, S., and Leahy, C.
\newblock The pile: An 800gb dataset of diverse text for language modeling.
\newblock \emph{CoRR}, abs/2101.00027, 2021.

\bibitem[Han et~al.(2016)Han, Mao, and Dally]{DeepCompress}
Han, S., Mao, H., and Dally, W.~J.
\newblock Deep compression: Compressing deep neural network with pruning,
  trained quantization and huffman coding.
\newblock In Bengio, Y. and LeCun, Y. (eds.), \emph{{ICLR}}, 2016.

\bibitem[Kim et~al.(2023)Kim, Lee, Kim, Park, Yoo, Kwon, and Lee]{PEQA}
Kim, J., Lee, J.~H., Kim, S., Park, J., Yoo, K.~M., Kwon, S.~J., and Lee, D.
\newblock Memory-efficient fine-tuning of compressed large language models via
  sub-4-bit integer quantization.
\newblock In \emph{NeurIPS}, pp.\  36187--36207, 2023.

\bibitem[Kim et~al.(2024)Kim, Hooper, Gholami, Dong, Li, Shen, Mahoney, and
  Keutzer]{SqueezeLLM}
Kim, S., Hooper, C., Gholami, A., Dong, Z., Li, X., Shen, S., Mahoney, M.~W.,
  and Keutzer, K.
\newblock Squeezellm: Dense-and-sparse quantization.
\newblock In \emph{ICML}, 2024.

\bibitem[Lee et~al.(2024)Lee, Jin, Kim, Kim, and Park]{OWQ}
Lee, C., Jin, J., Kim, T., Kim, H., and Park, E.
\newblock {OWQ:} lessons learned from activation outliers for weight
  quantization in large language models.
\newblock In \emph{{AAAI}}, 2024.

\bibitem[Li et~al.(2021)Li, Gong, Tan, Yang, Hu, Zhang, Yu, Wang, and
  Gu]{BRECQ}
Li, Y., Gong, R., Tan, X., Yang, Y., Hu, P., Zhang, Q., Yu, F., Wang, W., and
  Gu, S.
\newblock {BRECQ:} pushing the limit of post-training quantization by block
  reconstruction.
\newblock In \emph{{ICLR}}, 2021.

\bibitem[Li et~al.(2024)Li, Yu, Liang, He, Karampatziakis, Chen, and
  Zhao]{LoftQ}
Li, Y., Yu, Y., Liang, C., He, P., Karampatziakis, N., Chen, W., and Zhao, T.
\newblock Loftq: Lora-fine-tuning-aware quantization for large language models.
\newblock In \emph{{ICLR}}, 2024.

\bibitem[Lin et~al.(2024)Lin, Tang, Tang, Yang, Chen, Wang, Xiao, Dang, Gan,
  and Han]{AWQ}
Lin, J., Tang, J., Tang, H., Yang, S., Chen, W.-M., Wang, W.-C., Xiao, G.,
  Dang, X., Gan, C., and Han, S.
\newblock Awq: Activation-aware weight quantization for llm compression and
  acceleration.
\newblock In \emph{MLSys}, 2024.

\bibitem[Liu et~al.(2022{\natexlab{a}})Liu, Li, Wu, and Lee]{LLaVA}
Liu, H., Li, C., Wu, Q., and Lee, Y.~J.
\newblock Visual instruction tuning.
\newblock In \emph{NeurIPS}, pp.\  34892--34916, 2022{\natexlab{a}}.

\bibitem[Liu et~al.(2022{\natexlab{b}})Liu, Cheng, Huang, Xing, and Shen]{N2UQ}
Liu, Z., Cheng, K., Huang, D., Xing, E.~P., and Shen, Z.
\newblock Nonuniform-to-uniform quantization: Towards accurate quantization via
  generalized straight-through estimation.
\newblock In \emph{{CVPR}}, pp.\  4932--4942, 2022{\natexlab{b}}.

\bibitem[Loshchilov \& Hutter(2017)Loshchilov and Hutter]{Cosine}
Loshchilov, I. and Hutter, F.
\newblock {SGDR:} stochastic gradient descent with warm restarts.
\newblock In \emph{{ICLR}}, 2017.

\bibitem[Loshchilov \& Hutter(2019)Loshchilov and Hutter]{ADAMW}
Loshchilov, I. and Hutter, F.
\newblock Decoupled weight decay regularization.
\newblock In \emph{{ICLR}}, 2019.

\bibitem[Merity et~al.(2017)Merity, Xiong, Bradbury, and Socher]{wikitext}
Merity, S., Xiong, C., Bradbury, J., and Socher, R.
\newblock Pointer sentinel mixture models.
\newblock In \emph{{ICLR}}, 2017.

\bibitem[Oh et~al.(2022)Oh, Sim, Kim, and Lee]{SubsetQ}
Oh, S., Sim, H., Kim, J., and Lee, J.
\newblock Non-uniform step size quantization for accurate post-training
  quantization.
\newblock In \emph{{ECCV}}, pp.\  658--673, 2022.

\bibitem[Paszke et~al.(2019)Paszke, Gross, Massa, Lerer, Bradbury, Chanan,
  Killeen, Lin, Gimelshein, Antiga, Desmaison, K{\"{o}}pf, Yang, DeVito,
  Raison, Tejani, Chilamkurthy, Steiner, Fang, Bai, and Chintala]{PyTorch}
Paszke, A., Gross, S., Massa, F., Lerer, A., Bradbury, J., Chanan, G., Killeen,
  T., Lin, Z., Gimelshein, N., Antiga, L., Desmaison, A., K{\"{o}}pf, A., Yang,
  E.~Z., DeVito, Z., Raison, M., Tejani, A., Chilamkurthy, S., Steiner, B.,
  Fang, L., Bai, J., and Chintala, S.
\newblock Pytorch: An imperative style, high-performance deep learning library.
\newblock In \emph{NeurIPS}, pp.\  8024--8035, 2019.

\bibitem[Raffel et~al.(2020)Raffel, Shazeer, Roberts, Lee, Narang, Matena,
  Zhou, Li, and Liu]{C4}
Raffel, C., Shazeer, N., Roberts, A., Lee, K., Narang, S., Matena, M., Zhou,
  Y., Li, W., and Liu, P.~J.
\newblock Exploring the limits of transfer learning with a unified text-to-text
  transformer.
\newblock \emph{JMLR}, 21:\penalty0 140:1--140:67, 2020.

\bibitem[Sakaguchi et~al.(2020)Sakaguchi, Bras, Bhagavatula, and
  Choi]{WinoGrande}
Sakaguchi, K., Bras, R.~L., Bhagavatula, C., and Choi, Y.
\newblock Winogrande: An adversarial winograd schema challenge at scale.
\newblock In \emph{AAAI}, pp.\  8732--8740, 2020.

\bibitem[Shao et~al.(2024)Shao, Chen, Zhang, Xu, Zhao, Li, Zhang, Gao, Qiao,
  and Luo]{OmniQuant}
Shao, W., Chen, M., Zhang, Z., Xu, P., Zhao, L., Li, Z., Zhang, K., Gao, P.,
  Qiao, Y., and Luo, P.
\newblock Omniquant: Omnidirectionally calibrated quantization for large
  language models.
\newblock In \emph{ICLR}, 2024.

\bibitem[Touvron et~al.(2023)Touvron, Lavril, Izacard, Martinet, Lachaux,
  Lacroix, Rozi{\`{e}}re, Goyal, Hambro, Azhar, Rodriguez, Joulin, Grave, and
  Lample]{LLaMA}
Touvron, H., Lavril, T., Izacard, G., Martinet, X., Lachaux, M., Lacroix, T.,
  Rozi{\`{e}}re, B., Goyal, N., Hambro, E., Azhar, F., Rodriguez, A., Joulin,
  A., Grave, E., and Lample, G.
\newblock Llama: Open and efficient foundation language models.
\newblock \emph{CoRR}, abs/2302.13971, 2023.

\bibitem[Vaswani et~al.(2017)Vaswani, Shazeer, Parmar, Uszkoreit, Jones, Gomez,
  Kaiser, and Polosukhin]{Transformer}
Vaswani, A., Shazeer, N., Parmar, N., Uszkoreit, J., Jones, L., Gomez, A.~N.,
  Kaiser, L., and Polosukhin, I.
\newblock Attention is all you need.
\newblock In \emph{NeurIPS}, pp.\  5998--6008, 2017.

\bibitem[Wei et~al.(2022)Wei, Gong, Li, Liu, and Yu]{QDROP}
Wei, X., Gong, R., Li, Y., Liu, X., and Yu, F.
\newblock Qdrop: Randomly dropping quantization for extremely low-bit
  post-training quantization.
\newblock In \emph{{ICLR}}, 2022.

\bibitem[Wei et~al.(2023)Wei, Zhang, Li, Zhang, Gong, Guo, and
  Liu]{Suppression+}
Wei, X., Zhang, Y., Li, Y., Zhang, X., Gong, R., Guo, J., and Liu, X.
\newblock Outlier suppression+: Accurate quantization of large language models
  by equivalent and optimal shifting and scaling.
\newblock \emph{CoRR}, abs/2304.09145, 2023.

\bibitem[Xiao et~al.(2023)Xiao, Lin, Seznec, Wu, Demouth, and Han]{SmoothQuant}
Xiao, G., Lin, J., Seznec, M., Wu, H., Demouth, J., and Han, S.
\newblock Smoothquant: Accurate and efficient post-training quantization for
  large language models.
\newblock In \emph{{ICML}}, pp.\  38087--38099, 2023.

\bibitem[Xu et~al.(2024)Xu, Xie, Gu, Chen, Chang, Zhang, Chen, Zhang, and
  Tian]{QA-LoRA}
Xu, Y., Xie, L., Gu, X., Chen, X., Chang, H., Zhang, H., Chen, Z., Zhang, X.,
  and Tian, Q.
\newblock Qa-lora: Quantization-aware low-rank adaptation of large language
  models.
\newblock In \emph{{ICLR}}, 2024.

\bibitem[Yao et~al.(2022)Yao, Aminabadi, Zhang, Wu, Li, and He]{ZeroQuant}
Yao, Z., Aminabadi, R.~Y., Zhang, M., Wu, X., Li, C., and He, Y.
\newblock Zeroquant: Efficient and affordable post-training quantization for
  large-scale transformers.
\newblock In \emph{NeurIPS}, pp.\  27168--27183, 2022.

\bibitem[Yuan et~al.(2023)Yuan, Niu, Liu, Liu, Wang, Shang, Sun, Wu, Wu, and
  Wu]{RPTQ}
Yuan, Z., Niu, L., Liu, J., Liu, W., Wang, X., Shang, Y., Sun, G., Wu, Q., Wu,
  J., and Wu, B.
\newblock {RPTQ:} reorder-based post-training quantization for large language
  models.
\newblock \emph{CoRR}, abs/2304.01089, 2023.

\bibitem[Zeng et~al.(2023)Zeng, Liu, Du, Wang, Lai, Ding, Yang, Xu, Zheng, Xia,
  Tam, Ma, Xue, Zhai, Chen, Liu, Zhang, Dong, and Tang]{GLM}
Zeng, A., Liu, X., Du, Z., Wang, Z., Lai, H., Ding, M., Yang, Z., Xu, Y.,
  Zheng, W., Xia, X., Tam, W.~L., Ma, Z., Xue, Y., Zhai, J., Chen, W., Liu, Z.,
  Zhang, P., Dong, Y., and Tang, J.
\newblock {GLM-130B:} an open bilingual pre-trained model.
\newblock In \emph{{ICLR}}, 2023.

\bibitem[Zhang et~al.(2022)Zhang, Roller, Goyal, Artetxe, Chen, Chen, Dewan,
  Diab, Li, Lin, Mihaylov, Ott, Shleifer, Shuster, Simig, Koura, Sridhar, Wang,
  and Zettlemoyer]{OPT}
Zhang, S., Roller, S., Goyal, N., Artetxe, M., Chen, M., Chen, S., Dewan, C.,
  Diab, M.~T., Li, X., Lin, X.~V., Mihaylov, T., Ott, M., Shleifer, S.,
  Shuster, K., Simig, D., Koura, P.~S., Sridhar, A., Wang, T., and Zettlemoyer,
  L.
\newblock {OPT:} open pre-trained transformer language models.
\newblock \emph{CoRR}, abs/2205.01068, 2022.

\end{thebibliography}
\bibliographystyle{icml2025}

\newpage
\appendix
\onecolumn


\end{document}